# 'Say EM' for Selecting Probabilistic Models for Logical Sequences

**Kristian Kersting**
Machine Learning Laboratory
University of Freiburg
Georges-Koehler-Allee 079, 79112 Freiburg, Germany

**Tapani Raiko**
Laboratory of Computer and Information Science
Helsinki University of Technology
P.O. Box 5400, 02015 HUT, Finland

## Abstract

Many real world sequences such as protein secondary structures or shell logs exhibit a rich internal structures. Traditional probabilistic models of sequences, however, consider sequences of flat symbols only. *Logical hidden Markov models* have been proposed as one solution. They deal with logical sequences, i.e., sequences over an alphabet of logical atoms. This comes at the expense of a more complex model selection problem. Indeed, different abstraction levels have to be explored. In this paper, we propose a novel method for selecting logical hidden Markov models from data called SAGEM. SAGEM combines *generalized expectation maximization*, which optimizes parameters, with structure search for model selection using *inductive logic programming* refinement operators. We provide convergence and experimental results that show SAGEM's effectiveness.

## 1 Introduction

Hidden Markov models [21] (HMMs) are extremely popular for analyzing sequential data. Areas of application include computational biology, user modeling, and robotics. Despite their successes, HMMs have a major weakness: they handle only sequences of flat, i.e., unstructured symbols. In many applications the symbols occurring in sequences are structured. Consider, e.g., the sequence of UNIX commands `emacs lohmms.tex, ls, latex lohmms.tex, ...` Such data have been used to train HMMs for anomaly detection [15]. However, as the above command sequence shows, UNIX commands may have parameters. Thus, commands are essentially *structured* symbols. HMMs cannot easily deal with this type of structured sequences. Typically, the application of HMMs requires either 1) ignoring the structure of the commands (i.e., the parameters), or 2) taking all possible parameters explicitly into account. The former approach results in serious information loss; the latter in a combinatorial explosion in the number of parameters and, as a consequence, inhibits generalization.

The above sketched problem with HMMs is akin to the problem of dealing with structured examples in traditional machine learning algorithms as studied in the field of inductive logic programming (ILP) [17]. Recently, Kersting et al. [12] proposed *logical HMMs* (LOHMMs) as an *probabilistic* ILP [4] framework that upgrades HMMs to deal with structure. The key idea is to employ logical atoms. Using logical atoms, the above UNIX command sequence can be represented as `emacs(lohmms.tex), ls, latex(lohmms.tex), ...` LOHMMs have been proven to be useful within bioinformatics domains. For instance in [12], the LOHMMs used to discover structural signatures of protein folds were simpler but more effective compared to corresponding HMMs (120 vs. $> 62000$ parameters). The compactness and comprehensibility, however, comes at the expense of a more complex model selection problem. So far, model selection for LOHMMs has not been investigated. Our main contribution is SAGEM, German for 'say EM', a novel method for selecting LOHMM structures from data. Selecting a structure is a significant problem for many reasons. First, eliciting LOHMMs from experts can be a laborious and expensive process. Second, HMMs are commonly learned by estimating the maximum likelihood parameters of a fixed, fully connected model. Such an approach is not feasible for LOHMMs as different abstraction levels have to be explored. Third, LOHMMs are strictly more expressive than HMMs. In [11], LOHMMs are used to classify tree-structured mRNA data. Finally, the parameter estimation of a LOHMM is a costly nonlinear optimization problem, so the naïve search is infeasible.

SAGEM adapts Friedman's *structural EM* [6]. It combines a *generalized expectation maximization* (GEM) algorithm, which optimizes parameters, with structure search for model selection using ILP refinement operators. Thus, SAGEM explores different abstraction levels due to ILP re-finement operators, and, due to a GEM approach, it reduces the selection problem to a more efficiently solvable one.

The outline of the paper is as follows. Section 2 reviews LOHMMs and their underlying logical concepts; Section 3 formalizes the model selection problem; in Section 4, we present a naïve learning algorithm; in Section 5, we introduce a <u>s</u>tructur<u>a</u>l, <u>g</u>eneralized <u>EM</u> – called SAGEM – for learning LOHMMs. SAGEM is experimentally evaluated in Section 6. Before concluding we discuss related work.

## 2 Probabilistic Models for Logical Sequences

We will briefly review *logical Hidden Markov models* (LOHMMs) [12, 13, 11]. The logical component of HMMs corresponds to a *Mealy machine*, i.e., to a finite state machine where the output symbols are associated with transitions. The key idea to develop probabilistic models for structured sequences is to replace these flat symbols by abstract symbols, more precisely logical atoms.

**First-Order Predicate Logic:** A *first-order logic alphabet* $\Sigma$ is a set of relation symbols r with arity $m \geq 0$, written r/m, and a set of function symbols f with arity $n \geq 0$, written f/n. An *atom* $r(t_1, \ldots, t_m)$ is a relation symbol r followed by a bracketed $m$-tuple of terms $t_i$. A *term* is a variable V or a function symbol f of arity $n$ immediately followed by a bracketed $n$-tuple of terms $s_j$, i.e., $f(s_1, \ldots, s_n)$. A *definite clause* $A \leftarrow B$ consists of atoms $A$ and $B$ and can be read as A *is true if* B *is true*. A substitution $\theta = \{V_1/t_1, \ldots, V_k/t_k\}$, e.g. $\{X/\text{tex}\}$, is an assignment of terms $t_i$ to variables $V_i$. Applying a substitution $\theta$ to a term, atom or clause $e$ yields the instantiated term, atom, or clause $e\theta$ where all occurrences of the variables $V_i$ are simultaneously replaced by the term $t_i$, e.g. $ls(X) \leftarrow emacs(F,X)\{X/tex\}$ yields $ls(tex) \leftarrow emacs(F,tex)$. A term, atom or clause $e$ is called *ground* when it contains no variables, i.e., $vars(e) = \emptyset$. The *Herbrand base* of $\Sigma$, denoted as $hb_\Sigma$, is the set of all ground atoms constructed with the predicate and function symbols in $\Sigma$. The set $G_\Sigma(A)$ of an atom A consists of all ground atoms $A\theta$ belonging to $hb_\Sigma$.

Our running example will be user modeling. For example, $emacs(readme, other)$ means that the user of type $other$ writes a command $emacs\ readme$ to a shell.

**Logical Hidden Markov Models (LOHMMs):** The sequences generated by LOHMMs are sequences of ground atoms rather than flat symbols. Within LOHMMs, the flat symbols employed in traditional HMMs are replaced by logical atoms such as $emacs(F,tex)$. Each atom $emacs(F,tex)$ there represents the set of ground atoms $G_\Sigma(emacs(F,tex))$, e.g. $emacs(readme,tex) \in G_\Sigma(emacs(F,tex))$.

Additionally, we assume that the alphabet is typed which in our case means that there is a function mapping every predicate r/m and number $1 \leq i \leq m$ to the set of ground terms allowed as the $i$-th argument of predicate r/m. This set is called the domain of the $i$-th argument of predicate r/m.

Figure 1 shows a LOHMM graphically. The states, observations, and transitions of LOHMMs are **abstract** in the sense that every abstract state or observation A represents all possible concrete states in $G_\Sigma(A)$. In Figure 1 *solid edges* encode **abstract transitions**. Let H and B be logical atoms representing abstract states, let O be a logical atom representing an abstract output symbol. An abstract transition from state B with probability $p$ to state H and omitting O is denoted by $p : H \xleftarrow{O} B$. If H, B, and O are all ground, there is no difference to 'normal' transitions. Otherwise, if H, B, and O have no variables in common, the only difference to 'normal' transitions is that for each abstract state (resp. observation) we have to sample which concrete state (resp. observation) we are in. Otherwise, we have to remember the variable bindings. More formally, let $B\theta_B \in G_\Sigma(B)$, $H\theta_B\theta_H \in G_\Sigma(H\theta_B)$ $O\theta_B\theta_H\theta_O \in G_\Sigma(O\theta_B\theta_H)$, and let $\mu$ be a **selection distribution**. Then with probability $p \cdot \mu(H\theta_B\theta_H \mid H\theta_B) \cdot \mu(O\theta_B\theta_H\theta_O \mid O\theta_B\theta_H)$, the model makes a transition from state $B\theta_B$ to $H\theta_B\theta_H$ and emits symbol $O\theta_B\theta_H\theta_O$.

A selection distribution specifies for each abstract state (respectively observation) A over the alphabet $\Sigma$ a distribution $\mu(\cdot \mid A)$ over $G_\Sigma(A)$. Consider, for example, the abstract transition $0.5 : s(f(Z)) \xleftarrow{o(X,Y,Z)} s(X)$. Suppose, $B\theta_B = s(1)$, $\mu(s(f(3)) \mid s(f(Z))) = 0.2$ and $\mu(o(1,2,3) \mid o(1,Y,3)) = 0.05$. Then, from state $s(1)$ with probability $0.5 \times 0.2 \times 0.05 = 0.005$ the output symbol is $o(1,2,3)$ and the next state is $s(f(3))$. To reduce the model complexity, we employ a naïve Bayes approach in which – at the expense of a lower expressivity – functors are neglected and variables are treated independently. More precisely, for each domain $D_i$ there is a probability distribution $P_{D_i}$. Let $vars(A) = \{V_1, \ldots, V_l\}$ be the variables occurring in A, and let $\theta = \{t_1/V_1, \ldots t_l/V_l\}$ be a substitution grounding A. Each $V_j$ is then considered a random variable over the domain of the first argument of r/m it appears in, denoted by $D_{V_j}$. Then, $\mu(A\theta \mid A) = \prod_{j=1}^{l} P_{D_{V_j}}(V_j = t_j)$. For instance, $\mu(s(f(3)) \mid s(f(Z)))$ equals $P_1^{s/1}(3)$.

Indeed, multiple abstract transitions can match a given ground state. Consider the abstract states $B_1 = emacs(File, tex)$ and $B_2 = emacs(File, User)$ in Fig. 1 **(a)**. The abstract state $B_1$ is more specific than $B_2$ because there exists a substitution $\theta = \{User/tex\}$ such that $B_2\theta = B_1$, i.e., $B_2$ subsumes $B_1$. Therefore $G_\Sigma(B_1) \subseteq G_\Sigma(B_2)$ and the first transition can be regarded as more informative than the second one. It should therefore be preferred over the second one when starting e.g. from emacs(hmm1,tex). We will also say that the *transitions* of the first abstract state are *more specific* than the second ones; encoded by *dashed edges*. These considerations lead to the **conflict resolution strategy** [^1] of only considering the maximally specific transitions that apply to a state in order to determine the successor states. This implements a kind of exception handling or default reasoning and is akin to Katz's *back-off* $n$-gram models [10]. In back-off $n$-gram models, the most detailed model that is deemed to provide sufficiently reliable information about the current context is used. That is, if one encounters an $n$-gram that is not sufficiently reliable, then back-off to use an $(n-1)$-gram; if that is not reliable either then back-off to level $n-2$, etc.

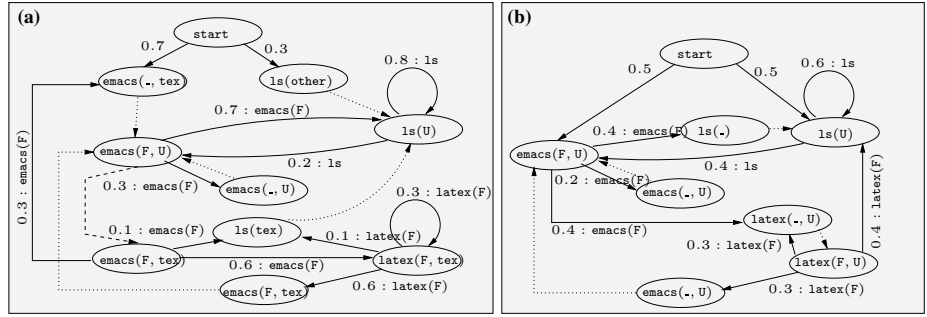

Figure 1: Logical hidden Markov models. The vertices represent abstract (hidden) states. Solid edges encode abstract transitions. Dotted edges indicate that two abstract states behave in exactly the same way. Dashed edge denote the *more general than* relation. The LOHMMs are described in the text.

Finally, *dotted edges* denote that two abstract states behave in exactly the same way. If we follow a transition to an abstract state with an outgoing dotted edge, we will automatically follow that edge making appropriate unifications.

**Definition 1** *A logical hidden Markov model (LOHMM) is a tuple $M = (\Sigma, \mu, \Delta)$ where $\Sigma$ is a logical alphabet, $\mu$ a selection probability over $\Sigma$ and $\Delta$ is a set of abstract transitions. Let $\mathbf{B}$ be the set of all atoms that occur as the body part of transitions in $\Delta$. We require*

$$\forall \mathtt{B} \in \mathbf{B} : \sum_{p : \mathtt{H} \xleftarrow{\mathtt{O}} \mathtt{B} \in \Delta} p = 1. \quad (1)$$

In [11] it is proven that LOHMMs specify a unique probability measure over $hb_\Sigma$. Here, we would like to exemplify that LOHMMs are generative models. Consider the model in Fig. 1(**a**). Starting from start, it chooses an initial abstract state, say emacs(_,tex) with probability 0.7. Here, _ denotes an anonymous variable which is read and treated as distinct, new variables each time it is encountered. Forced to follow the dotted edge, it enters the abstract state emacs(F,U). In each abstract state, the model samples values for all variables that are not instantiated yet according to the *selection distribution* $\mu$. Since the value of U was already instantiated in the previous abstract state emacs(F,tex), the model has only to sample a value for F, say f1, using $\mu$. Now, it selects a transition, say, to latex(F,tex) with probability 0.6. Since F is shared among the head and the body, the state latex(f1,tex) is selected with probability 1.0. The observation emacs(f1) is emitted from emacs(F) with probability 1.0 using $\mu$. Now, the model goes over to, say ls(tex), emitting latex(f1) which in turn was sampled from latex(F). The dotted edge brings us to ls(U) and automatically unifies U with tex. Emitting ls, we return to emacs(F,tex) where F now denotes a new filename.

## 3 The Learning Setting

For traditional HMMs, the learning problem basically collapses to parameter estimation (i.e., estimating the transition probabilities) because HMMs can be considered to be fully connected. For LOHMMs, however, we have to account for different abstraction levels. The model selection problem can formally be defined as:

**Given** a set $\mathbf{O} = \{O_1, \ldots, O_m\}$ of data cases independently sampled from the same distribution, a set $\mathcal{M}$ of LOHMMs, and a scoring function $score_\mathbf{O} : \mathcal{M} \mapsto \mathbb{R}$, **find** a hypothesis $M^* \in \mathcal{M}$ that maximizes $score_\mathbf{O}$.

Each *data case* $O_i \in \mathbf{O}$ is a sequence $O_i = \mathtt{o_{i,1}o_{i,2} \ldots o_{i,T}}$ of ground atoms and describes the observations evolving over time. For instance in the user modeling domain a data case could be emacs(lohmms), ls, emacs(lohmms). The corresponding evolution of the system's state over time $H_i = \mathtt{h_{i,0}h_{i,1} \ldots h_{i,T_i+1}}$ is hidden, i.e., not specified in $O_i$. For instance, we do not know whether emacs(lohmms) has been generated by emacs(lohmms, prog) or emacs(lohmms, tex).

The *hypothesis space* $\mathcal{M}$ consists of all candidate LOHMMs to be considered during search. We assume $\Sigma$

[^1]: Another conflict resolution strategy would be *smoothing*, i.e., considering all matching abstract states. We chose not to use smoothing to keep the LOHMM locally interpretable, i.e. to have a single abstract body for each ground state.

to be given. Thus, the possible constants which can be selected by $\mu$ are apriori known. Each $M \in \mathcal{M}$ is parameterized by a vector $\boldsymbol{\lambda}_M$. Each (legal) choice of $\boldsymbol{\lambda}_M$ defines a probability distribution $P(\cdot \mid M, \boldsymbol{\lambda}_M)$ over $\text{hb}(\Sigma)$. For the sake of simplicity, we will denote the underlying logic program (i.e., the set of abstract transitions without associated probability values) by $M$ and abbreviate $\boldsymbol{\lambda}_M$ by $\boldsymbol{\lambda}$ as long as no ambiguities will arise. Furthermore, a syntactic bias on the transitions to be induced is a parameter of our framework, as usual in ILP [18]. For instance in the experiments, we only consider transitions which obey the type constraints induced by the predicates.

As score, we employ $score_{\mathbf{O}}(M, \boldsymbol{\lambda}) = \log P(\mathbf{O} \mid M, \boldsymbol{\lambda}) - Pen(M, \boldsymbol{\lambda}, \mathbf{O})$. Here, $\log P(\mathbf{O} \mid M, \boldsymbol{\lambda})$ is the *log-likelihood* of the current of model $(M, \boldsymbol{\lambda})$. It holds that the higher the log-likelihood, the closer $(M, \boldsymbol{\lambda})$ models the probability distribution induced by the data. The second term, $Pen(M, \boldsymbol{\lambda}, \mathbf{O})$, is a penalty function that biases the scoring function to prefer simpler models. Motivated by the *minimum description length* score for Bayesian networks, we use the simple penalty $Pen(M, \boldsymbol{\lambda}, \mathbf{O}) = |\Delta| \log(m)/2$. It is independent of the model parameters and therefore it can be neglected when estimating parameters. We assume that each $M$ covers all possible observation sequences (over the given language $\Sigma$). This guarantees that all new data cases will get a positive likelihood.

## 4 A Naïve Learning Algorithm

A simple way of selecting a model structure is the following greedy approach:

1:    Let $\boldsymbol{\lambda}^0 = \operatorname{argmax}_{\boldsymbol{\lambda}} score_{\mathbf{O}}(M^0, \boldsymbol{\lambda})$
2:    **Loop** for $k = 0, 1, 2, \ldots$
3:       **Find** model $M^{k+1} \in \rho(M^k)$ that maximizes
           $\max_{\boldsymbol{\lambda}} score_{\mathbf{O}}(M^{k+1}, \boldsymbol{\lambda})$
4:       Let $\boldsymbol{\lambda}^{k+1} = \operatorname{argmax}_{\boldsymbol{\lambda}} score_{\mathbf{O}}(M^{k+1}, \boldsymbol{\lambda})$
5:    **Until** convergence, i.e., no improvement in score

It takes as input an initial model $M^0$ and the data $\mathbf{O}$. At each stage $k$ we choose a model structure and parameters among the current best model $M^k$ and its neighbors $\rho(M^k)$ (see below) that have the highest score. It stops, when there is no improvement in score. In practice, we initialize the parameters of each model on lines 1 and 3 randomly.

We will now show how to traverse the hypotheses space and how to estimate parameters for a hypothesis in order to score it. That is, we will make line 3 more concrete.

**Traversing the Hypotheses Space:** An obvious candidate for the initial hypothesis $M^0$ (which we also used in our experiments) is the fully connected LOHMM built over all maximally general atoms over $\Sigma$, i.e., expressions of the form $\text{r}(\text{X}_1, ..., \text{X}_\text{m})$, where the $\text{X}_\text{i}$ are different variables.

Now, to traverse the hypothesis space $\mathcal{M}$, we have to compute all neighbors of the currently best hypothesis $M^k$. To do so, we employ refinement operators traditionally used in ILP. More precisely, for the language bias considered and the experiments conducted in the present paper, we used the refinement operator $\rho : \mathcal{M} \mapsto 2^{\mathcal{M}}$ which selects a single clause $\text{cl} \equiv p : \text{H} \xleftarrow{\text{O}} \text{B} \in \mathcal{M}$ and adds a minimal specialization $\text{cl}' \equiv p : \text{H}' \xleftarrow{\text{O}'} \text{B}'$ of cl to $\mathcal{M}$ (w.r.t. to $\theta$-subsumption). Specializing a single abstract transition means instantiating or unifying variables, i.e., $\text{cl}' \equiv \text{cl}\,\theta$ for some substitution $\theta$. When adding $\text{cl}'$ to $M^k$, we have to ensure that (1) the same observation and hidden state sequences are still covered and (2) the list of bodies $\mathbf{B}'$ after applying $\rho(M)$ should remain well-founded, that is, for each ground state, there is a unique maximally specific body in $\mathbf{B}'$. Both conditions together guarantee that the most specific body corresponding to a state always exists and is unique. Condition (1) can only be violated if $\text{B}' \notin \mathbf{B}$. In this case, we add transitions with $\text{B}'$ and maximally general heads and observations. Condition (2) is established analogously. We complete the keep the list of bodies well-founded by adding new bodies (and therefore abstract transitions) in a similar way as described above.

Consider refining the LOHMM in Fig. 1 **(b)**. When adding $\text{ls}(\text{U}) \xleftarrow{\text{latex(lohmm)}} \text{latex}(\text{lohmm}, \text{U})$, hence introducing the more specific abstract state $\text{latex}(\text{lohmm}, \text{U})$, further variants of the same abstract transition but with different heads have to be added. Otherwise condition (1) would be violated as the resulting LOHMM does not cover the same sequences as the original one; the state $\text{latex}(\text{lohmm}, \text{U})$ can only be left via $\text{ls}(\text{U})$ and not e.g. via $\text{emacs}(\_, \text{U})$. On the other hand, we have to be careful when subsequently adding abstract transitions for the body $\text{latex}(\text{F}, \text{tex})$. The problem is that we do not know which abstract body to select in state $\text{latex}(\text{lohmm}, \text{tex})$. To fulfill condition (2), you need to add abstract transitions for an additional, third abstract state $\text{latex}(\text{lohmm}, \text{tex})$, too.

**Parameter Estimation:** In the presence of hidden variables maximum log-likelihood (ML) parameter estimation is a numerical optimization problem, and all known algorithms involve nonlinear, iterative optimization and multiple calls to an inference algorithm. The most common approach for HMMs is the Baum-Welch algorithm, an instance of the EM algorithm [5]. In each iteration $l + 1$ it performs two steps:

**(E-step)** *Compute the expectation of the log-likelihood given the old model $(M^k, \boldsymbol{\lambda}^{k,l})$ and the observed data $\mathbf{O}$, i.e., $Q(M^k, \boldsymbol{\lambda} \mid M^k, \boldsymbol{\lambda}^{k,l}) = E\left[\log P(\mathbf{O}, \mathbf{H} \mid M^k, \boldsymbol{\lambda}) \mid M^k, \boldsymbol{\lambda}^{k,l}\right]$.*

Here, $\mathbf{O}, \mathbf{H}$ denotes the completion of $\mathbf{O}$ where the evolution $\mathbf{H}$ of the system's state over time is made explicit. The current model $(M^k, \boldsymbol{\lambda}^{k,l})$ and the observed data $\mathbf{O}$ give us the conditional distribution governing **H**, and $E[\cdot|\cdot]$ denotes the expectation over it. The function $Q$ is called the *expected score*.

**(M-step)** *Maximize the expected score $Q(M^k, \boldsymbol{\lambda} \mid M^k, \boldsymbol{\lambda}^{k,l})$ w.r.t. $\boldsymbol{\lambda}$, i.e., $\boldsymbol{\lambda}^{k,l+1} = \text{argmax}_{\boldsymbol{\lambda}} Q(M^k, \boldsymbol{\lambda}|M^k, \boldsymbol{\lambda}^{k,l})$.*

The naïve greedy algorithm can easily be instantiated using the EM. The problem, however, is its huge computational costs. To evaluate a single neighbor, the EM has to run for a reasonable number of iterations in order to get reliable ML estimates of $\boldsymbol{\lambda}^{k'}$. Each EM iteration requires a full LOHMM inference on all data cases. In total, the running time per neighbor evaluation is at least $\mathcal{O}(\#\textit{EM iterations} \cdot \textit{size of data})$.

## 5  SAGEM: Structural Generalized EM

To reduce the computational costs, SAGEM (German for *'say EM'*) adapts Friedman's *structural EM* (SEM) [6]. That is, we take our current model $(M^k, \boldsymbol{\lambda}^k)$ and run the EM algorithm for a while to get reasonably completed data. We then fix the completed data cases and use them to compute the ML parameters $\boldsymbol{\lambda}^{k'}$ of each neighbor $M^{k'}$. We choose the neighbor with the best improvement of the score as $(M^{k+1}, \boldsymbol{\lambda}^{k+1})$ and iterate. More formally, we have

1:  Initialize $\boldsymbol{\lambda}^{0,0}$ randomly
2:  **Loop** for $k = 0, 1, 2, \ldots$
3:     **Loop** for $l = 0, 1, 2, \ldots$
4:        Let $\boldsymbol{\lambda}^{k,l+1} = \text{argmax}_{\boldsymbol{\lambda}} Q(M^k, \boldsymbol{\lambda} \mid M^k, \boldsymbol{\lambda}^{k,l})$
5:     **Until** convergence **or** $l = l_{\max}$
6:     **Find** model $M^{k+1} \in \rho(M^k)$ that maximizes
      $\max_{\boldsymbol{\lambda}} Q(M^{k+1}, \boldsymbol{\lambda} \mid M^k, \boldsymbol{\lambda}^{k,l})$
7:     Let $\boldsymbol{\lambda}^{k+1,0} = \text{argmax}_{\boldsymbol{\lambda}} Q(M^{k+1}, \boldsymbol{\lambda} \mid M^k, \boldsymbol{\lambda}^{k,l})$
8:  **Until** convergence

The hypotheses space is traversed as described in Section 4, and again we stop if there is no improvement in score. The following theorem shows that even when the structure changes in between, improving the expected score $Q$ always improves the log-likelihood as well.

**Theorem 1** *If $Q(M, \boldsymbol{\lambda} \mid M^k, \boldsymbol{\lambda}^{k,l}) > Q(M^k, \boldsymbol{\lambda}^{k,l} \mid M^k, \boldsymbol{\lambda}^{k,l})$ holds, then $\log P(\mathbf{O} \mid M, \boldsymbol{\lambda}) > \log P(\mathbf{O} \mid M^k, \boldsymbol{\lambda}^{k,l})$ holds.*

The proof is a simple extension of the argumentation by [16, Section 3.2 ff.]. To apply the algorithm to selecting LOHMMs, we will now show how to choose the best neighbour [2] in line 6.

---
[2] In the following, we will omit some derivation steps due to space restriction. They can be found in [13]. Furthermore, for the sake of simplicity, we will not explicitly check that a transition is *maximally specific* for ground states.

Let $c(\mathtt{b}, \mathtt{h}, \mathtt{o})$ denote the number of times the systems proceeds from ground state $\mathtt{b}$ to ground state $\mathtt{h}$ emitting ground observation $\mathtt{o}$. The expected score in line 6 simplifies to

$$Q(M, \boldsymbol{\lambda}|M^k, \boldsymbol{\lambda}^{k,l}) \qquad (2)$$
$$= \sum_{\mathtt{b},\mathtt{h},\mathtt{o}} \underbrace{E\left[c(\mathtt{b}, \mathtt{h}, \mathtt{o}) \Big| M^k, \boldsymbol{\lambda}^{k,l}\right]}_{=:ec(\mathtt{b},\mathtt{h},\mathtt{o})} \cdot \log P(\mathtt{h}, \mathtt{o}|\mathtt{b}, M, \boldsymbol{\lambda}) .$$

The term $ec(\mathtt{b}, \mathtt{h}, \mathtt{o})$ in (2) denotes the expected counts of making a transition from ground state $\mathtt{b}$ to ground state $\mathtt{h}$ emitting ground observation $\mathtt{o}$. The expectation is taken according to $(M^k, \boldsymbol{\lambda}^{k,l})$.

An analytical solution, however, of the M-step in line 7 seems to be difficult. In HMMs, the updated transition probabilities are simply directly proportional to the expected number of times they are used. In LOHMMs, however, there is an ambiguity: multiple abstract transitions (with the same body), can match the same ground transition $(\mathtt{b}, \mathtt{h}, \mathtt{o})$. Using $ec$ as sufficient statistics makes the M step nontrivial. The solution is to improve (2) instead of maximizing it. Such an approach is called *generalized EM* [16]. To do so, we follow a gradient-based optimization technique. We iteratively compute the *gradient* $\nabla_{\boldsymbol{\lambda}}$ of (2) w.r.t. the parameters of a LOHMM, and, then, take a step in the direction of the gradient to the point $\boldsymbol{\lambda} + \delta \nabla_{\boldsymbol{\lambda}}$ where $\delta$ is the step-size.

For LOHMMs, the gradient w.r.t. (2) consists of partial derivatives w.r.t. abstract transition probabilities and to selection probabilities. Assume that $\lambda$ is the transition probability associated with some abstract transition $\mathtt{cl}$. Now, the partial derivative of (2) w.r.t. some parameter $\lambda$ is

$$\frac{\partial Q(M, \boldsymbol{\lambda} \mid M^k, \boldsymbol{\lambda}^{k,l})}{\partial \lambda}$$
$$= \sum_{\mathtt{b},\mathtt{h},\mathtt{o}} ec(\mathtt{b}, \mathtt{h}, \mathtt{o}) \cdot \frac{\partial \log P(\mathtt{h}, \mathtt{o} \mid \mathtt{b}, M, \boldsymbol{\lambda})}{\partial \lambda}$$
$$= \sum_{\mathtt{b},\mathtt{h},\mathtt{o}} \frac{ec(\mathtt{b}, \mathtt{h}, \mathtt{o})}{P(\mathtt{h}, \mathtt{o} \mid \mathtt{b}, M, \boldsymbol{\lambda})} \cdot \frac{\partial P(\mathtt{h}, \mathtt{o} \mid \mathtt{b}, M, \boldsymbol{\lambda})}{\partial \lambda} \qquad (3)$$

The partial derivative of $P(\mathtt{h}, \mathtt{o} \mid \mathtt{b}, M, \boldsymbol{\lambda})$ w.r.t. $\lambda$ can be computed as

$$\frac{P(\mathtt{h}, \mathtt{o} \mid \mathtt{b}, M, \boldsymbol{\lambda})}{\partial \lambda}$$
$$= \mu(\mathtt{h} \mid \text{head}(\text{cl})\theta_{\mathtt{H}}, M) \cdot \mu(\mathtt{o} \mid \text{obs}(\text{cl})\theta_{\mathtt{H}}\theta_{\mathtt{O}}, M) \qquad (4)$$

Substituting (4) back into (3) yields

$$\frac{\partial Q(M, \boldsymbol{\lambda} \mid M^k, \boldsymbol{\lambda}^{k,l})}{\partial \lambda}$$
$$= \sum_{\mathtt{b},\mathtt{h},\mathtt{o}} \left( \frac{ec(\mathtt{b}, \mathtt{h}, \mathtt{o})}{P(\mathtt{h}, \mathtt{o} \mid \mathtt{b}, M, \boldsymbol{\lambda})} \cdot \mu(\mathtt{h} \mid \text{head}(\text{cl})\theta_{\mathtt{H}}, M) \right.$$
$$\left. \cdot \mu(\mathtt{o} \mid \text{obs}(\text{cl})\theta_{\mathtt{H}}\theta_{\mathtt{O}}, M) \right) \qquad (5)$$

The selection probability follows a naïve Bayes approach. Therefore, one can show in a similar way as for transition probabilities that

$$\frac{\partial Q(M, \boldsymbol{\lambda} \mid M^k, \boldsymbol{\lambda}^{k,l})}{\partial \lambda}$$
$$= \sum_{\mathtt{b,h,o}} \Big( \frac{ec(\mathtt{b,h,o})}{P(\mathtt{h,o} \mid \mathtt{b}, M, \boldsymbol{\lambda})} \cdot \sum_{\mathrm{cl}} c(\lambda, \mathrm{cl}, \mathtt{b, h, o}) \cdot$$
$$\cdot P(\mathrm{cl} \mid M, \boldsymbol{\lambda}) \cdot \mu(\mathtt{h} \mid \mathrm{head(cl)} \theta_{\mathtt{H}}, M) \cdot$$
$$\cdot \mu(\mathtt{o} \mid \mathrm{obs(cl)} \theta_{\mathtt{H}} \theta_{\mathtt{O}}, M) \Big) \quad (6)$$

where $c(\lambda, \mathrm{cl}, \mathtt{b, h, o})$ is the number of times that the domain element associated with $\lambda$ is selected to ground $\mathrm{cl}$ w.r.t. $\mathtt{h}$ and $\mathtt{o}$.

In the problem at hand, the described method has to be modified to take into account that $\lambda \in [0,1]$ and that corresponding $\lambda$'s sum to $1.0$. A general solution, which we used in our experiments, is to reparameterize the problem so that the new parameters automatically respect the constraints no matter what their values are. To do so, we define the parameters $\boldsymbol{\beta}$ where $\beta_{ij} \in \mathbb{R}$ such that $\lambda_{ij} = \exp(\beta_{ij})/(\sum_l \exp(\beta_{il}))$. This enforces the constraints given above, and a local maximum w.r.t. $\boldsymbol{\beta}$ is also a local maximum w.r.t. $\boldsymbol{\lambda}$, and vice versa. The gradient w.r.t the $\beta_{ij}$'s can be found by computing the gradient w.r.t. the $\lambda_{ij}$'s and then deriving the gradient w.r.t. $\boldsymbol{\beta}$ using the chain rule.

**Discussion on SAGEM:** What do we gain from SAGEM over the naïve approach? The expected ground counts $ec(\mathtt{b,h,o})$ are used as the sufficient statistics to evaluate all the neighbors. Evaluating neighbors is thus now independent of the number and length of the data cases— a feature which is important for scaling up. More precisely, the running time per neighbor evaluation is basically $\mathcal{O}(\#\textit{Gradient iterations} \cdot \#\textit{Ground transitions})$ because SAGEM's gradient approach does not perform LOHMM inferences.

The greedy approach is not always enough. For instance, if two hidden states are equivalent, to make them effectively differ from each other, one needs to make them differ both in visiting probabilities of the state and in behavior in the state, possibly requiring two steps for any positive effect. Fixing the expected counts in SAGEM worsens the problem, since changes in visiting probabilities of states do not show up before a LOHMM inference is made. To overcome this, different search strategies, such as beam search, can be used: Instead of a current hypothesis, a fixed-size set of current hypotheses is considered, and their common neighborhood is searched for the next set.

To summarize, SAGEM explores different abstraction levels due to ILP refinement operators, and, due to a GEM approach, it reduces the neighborhood evaluation problem to one that is solvable more efficiently.

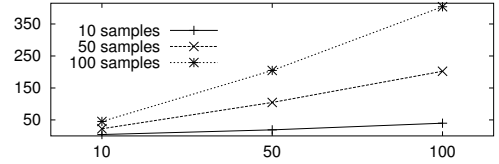

Figure 2: Speed-up (y axis), i.e., the ratio of time per EM iteration (in sec.) and time per SAGEM's gradient approach to evaluate neighbors. The speed-up is shown for different numbers of sequence lengths (x axis) and for different numbers of data cases (curves).

## 6 Experimental Evaluation

Our intentions here are to investigate whether SAGEM can be applied to real world domains. More precisely, we will investigate whether SAGEM

**H1** speeds-up neighbor evaluation considerably (compared to the naïve learning algorithm);
**H2** finds a comprehensible model;
**H3** works in the presence of transition ambiguity;
**H4** can be applied to real-world domains and is competitive with standard machine learning algorithms such as nearest-neighbor and decision-tree learners.

To this aim, we implemented the SAGEM using the Prolog system YAP-4.4.4. The experiments were run on a Pentium-III-2.3 GHz machine. For the improvement of expected score, we adapted the scaled conjugate gradient as implemented in Bishop and Nabney's Netlab library (http://www.ncrg.aston.ac.uk/netlab/) with a maximum number of $10$ iterations and $5$ random restarts.

**Experiments with Synthetic Data:** We sampled independently $10, 50, 100$ sequences of length $10, 50, 100$ ($100$ to $10000$ ground atoms in total) from the LOHMM shown in Fig. 1(**a**) and computed their ground counts w.r.t. the samples. We measured the averaged running time in seconds per iteration for both, the naïve algorithm and SAGEM's gradient approach to evaluate neighbors when applied to the LOHMM shown in Fig. 1(**b**). The times were measured using YAP's built-in statistics/2. The results are summarized in Fig. 2 showing the ratio of running times of naïve over SAGEM's gradient approach. In some cases the speed-up was more than $400$. EM's lowest running time was $0.075$ seconds (for $10$ sequences of length $10$). In contrast, SAGEM was constantly below $0.017$ seconds. This suggests that **H1** holds.

We sampled $2000$ sequences of length $15$ ($30000$ ground atoms) from the LOHMM in Fig. 1(**a**). There were $4$ filenames, $2$ users types. The initial hypothesis was the LOHMM in Fig. 1(**b**) with randomly initialized parameters. We run SAGEM on the sampled data. [3] Averaged over 5 runs, estimating the parameters for the initial hypothesis achieved a score of $-47203$. In contrast, the score of SAGEM's selected model was $-26974$ which was even slightly above the score of the original LOHMM ($-30521$). This suggests that **H3** holds. Moreover, in all runs, SAGEM included e.g. $\texttt{latex(A,B)} \xleftarrow{0.61:\texttt{emacs(A)}} \texttt{emacs(A,B)}$ and $\texttt{emacs(A,B)} \xleftarrow{0.48:\texttt{emacs(A)}} \texttt{latex(A,B)}$ which were not present in the initial model. This suggests that **H2** holds.

---

[3]The naïve algorithm was no longer used for comparison due

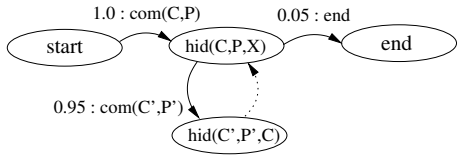

Figure 3: The initial hypothesis for the experiments with real-world data is a minimal structure, implying learning from scratch. $C$ stands for command and $P$ for parameters. The hidden state $hid$ contains the new command, parameters and the latest old command.

**Experiments with Real-World Data:** Finally, we applied SAGEM to the data set collected by Greenberg [8]. The data consists of 168 users of four groups: computer scientists, non-programmers, novices and others. About 300000 commands have been logged in on average 110 sessions per user. We present here results for two classes: novice-1(NV) consisting of 2512 ground atoms and non-programmers-4 (NP) consisting of 5183 ground atoms. We randomly selected 35 training sessions (about 1500 commands) for each class. On this data, we let SAGEM select a model for each class independently, starting from the initial hypothesis described in Fig. 3. To evaluate, we computed the plug-in estimates of each model for the remaining sessions corrected by the class priors. Averaged over five runs, the precision ($0.94 \pm 0.06$ NV, $0.91 \pm 0.02$ NP) and recall values ($0.67 \pm 0.03$ NV, $0.89 \pm 0.05$ NP) were balanced and the overall predictive accuracy was $0.92 \pm 0.01$. Jacobs and Blockeel [9] report that a kNN approach achieved a precision of $0.91$ and J48 (WEKA's implementation of Quinlan's C4.5 decision tree learner) of $0.86$ averaged over ten runs on 50 randomly sampled training examples. This suggests that **H4** holds. The used kNN and J48 methods, however, do not yield generative models and lack comprehensibility. SAGEM's selected models encoded e.g. "*non-programmers are very likely to type in* `cd..` *after performing* `ls` *in some directory*". This pattern was not present in the NV model. This suggests that **H2** holds.

## 7 Related Work

*Statistical relational learning* (SRL) can be viewed as combining ILP principles (such as refinement operators) with statistical learning, see [3] for an overview and references.

Most attention, however, has been devoted to developing highly expressive formalisms. LOHMMs can be seen as an attempt towards *downgrading* such highly expressive frameworks. They retain most of the essential logical features but are easier to understand, adapt and learn. For the same reasons, simple statistical techniques (such as logistic regression or naïve Bayes) have been combined with ILP refinement oprators for traversing the search space, see e.g. [20, 14]. They, however, do not select dynamic models.

LOHMMs are related to Anderson *et al.*'s *relational Markov models* (RMMs) [1]. Here, states can be of different types, with each type described by a different set of variables. The domain of each variable is hierarchically structured. The main differences are that neither variable bindings, unification nor hidden states are supported. RMMs do not select the most-specific transition to resolve conflicting transitions. Instead, they interpolate between conflicting ones. This is an interesting option for LOHMMs because it makes parameter estimation more robust. On the other hand, it also seems to make it more difficult to adhere one of our design principles: locally interpretable transitions. Structure learning has been addressed based on *probability estimation trees*.

Logical sequences can be converted into binary trees by putting each instance of a relation symbol into a node. The left subtree represents the first argument and the right subtree represents the next atom in the list (of observations or arguments). Methods for learning tree languages [2] can thus be used for learning probabilistic models for logical sequences, too. The main differences, though, is that variable bindings are not supported.

LOHMMs are related to several extensions of HMMs such as factorial HMMs [7]. Here, state variables are decomposed into smaller units. The key difference to LOHMMs is that these approaches do not employ logical concepts.

Finally, SAGEM is related to more advanced HMM model selection methods. *Model merging* [22] starts with the most specific model consistent with the training data and generalizes by successively merging states. Abstract transitions, however, aim at good generalization, and the most general clauses can be considered to be the most informative ones. Therefore, *successive state splitting* [19] refines hidden states by splitting them into new states. In both cases, the authors are not aware of adaptions of Friedman's SEM.

## 8 Conclusions

A novel model selection method for logical hidden Markov models called SAGEM has been introduced. SAGEM combines *generalized EM*, which optimizes parameters, with structure search for model selection using ILP refinement operators. Experiments show SAGEM's effectiveness.

Future work should address other scores; other refinement to unreasonable running times.

operators e.g. handling functors, deleting transitions, and generalizing hypotheses; logical pruning criteria for hypotheses; and efficient storing of ground counts. Moreover, the authors hope that the presented work will inspire further research at the intersection of ILP and HMMs.

**Acknowledgments:** The authors thank the three anonymous reviewers for helpful comments, Luc De Raedt for discussions on the topic, Nico Jacobs for discussions on user modeling, and Saul Greenberg for providing the data. The research was partially supported by the EU IST programme, contract no. FP6-508861, *APrIL II*, and by the Finnish Centre of Excellence Programme (2000-2005) under the project *New Information Processing Principles*. Tapani Raiko was partially supported by a Marie Curie fellowship at DAISY, HPMT-CT-2001-00251.